%% file: main.tex
\documentclass[conference]{IEEEtran}
\IEEEoverridecommandlockouts

\usepackage[utf8]{inputenc}
\usepackage[colorlinks,bookmarksopen,bookmarksnumbered,citecolor=black,urlcolor=black,linkcolor=black]{hyperref}
\usepackage{mathtools}
\usepackage{amsthm}
\usepackage{amssymb}
\usepackage{amsmath}
\usepackage{yhmath}
\usepackage{amsfonts}
\usepackage{url}
\usepackage{listings}
\usepackage{color}
\usepackage[table,xcdraw,dvipsnames]{xcolor}
\usepackage{graphicx}
\usepackage{caption}
\usepackage[noadjust]{cite}
\usepackage{breqn}
\usepackage{makecell}
\usepackage{enumitem}
\usepackage[titlenumbered,ruled,linesnumbered]{algorithm2e}
\usepackage{balance}
\usepackage{amssymb}
\usepackage{multirow}
\usepackage{multicol} 
\usepackage{soul}
\usepackage{subfigure}
\usepackage{adjustbox}
\usepackage{arydshln}
\usepackage{tikz}
\usepackage{booktabs}
\usepackage{makecell}
\usepackage{tcolorbox}
\usepackage[]{mdframed}
\tcbuselibrary{theorems}

\def\BibTeX{{\rm B\kern-.05em{\sc i\kern-.025em b}\kern-.08em
    T\kern-.1667em\lower.7ex\hbox{E}\kern-.125emX}}

\newtcbtheorem[]{characteristic}{Characteristic}{
    colback=green!5,
    colframe=green!35!black,
    fonttitle=\bfseries\small,
    fontupper=\small,
    boxrule=0.5mm,
    coltitle=white,
    width=\linewidth}
{def}

\newtcbtheorem[]{limitation}{Limitation}{
    colback=green!5,
    colframe=green!35!black,
    fonttitle=\bfseries\small,
    fontupper=\small,
    boxrule=0.5mm,
    width=\linewidth}
{def}

\definecolor{cmdcolor}{RGB}{34, 153, 84}
\definecolor{codegreen}{rgb}{0,0.6,0}
\definecolor{codegray}{rgb}{0.5,0.5,0.5}
\definecolor{codepurple}{rgb}{0.58,0,0.82}
\definecolor{backcolour}{rgb}{0.95,0.95,0.92}

\newcommand{\eg}{{\it e.g.}\xspace}
\newcommand{\ie}{{\it i.e.}\xspace}

\renewcommand{\arraystretch}{1.3}
\newcommand\lname{RSL}
\newcommand\fname{NRTrans}

\newcommand{\blackcircled}[1]{%
\tikz[baseline=(char.base)]{
\node[shape=circle, scale=0.8,draw=black,inner sep=0.5pt,fill=black, text=white] (char) {#1};}}

\begin{document}

\title{An LLM-powered Natural-to-Robotic Language Translation Framework with Correctness Guarantees}

\author{
\IEEEauthorblockN{1\textsuperscript{st} ZhenDong Chen}
\IEEEauthorblockA{\textit{School of Computer Science and Engineering} \\
\textit{Sun Yat-Sen University}\\
GuangZhou, China \\
chenzhd29@mail2.sysu.edu.cn} \\

\IEEEauthorblockN{3\textsuperscript{rd} ShiXing Wan}
\IEEEauthorblockA{\textit{School of Computer Science and Engineering} \\
\textit{Sun Yat-Sen University}\\
GuangZhou, China \\
wanshx3@mail2.sysu.edu.cn} \\

\IEEEauthorblockN{5\textsuperscript{th} YongTian Cheng}
\IEEEauthorblockA{\textit{School of Computer Science and Engineering} \\
\textit{Sun Yat-Sen University}\\
GuangZhou, China \\
chengyt8@mail2.sysu.edu.cn}
\and
\IEEEauthorblockN{2\textsuperscript{nd} ZhanShang Nie}
\IEEEauthorblockA{\textit{School of Computer Science and Engineering} \\
\textit{Sun Yat-Sen University}\\
GuangZhou, China \\
niezhsh@mail2.sysu.edu.cn} \\

\IEEEauthorblockN{4\textsuperscript{th} JunYi Li}
\IEEEauthorblockA{\textit{School of Computer Science and Engineering} \\
\textit{Sun Yat-Sen University}\\
GuangZhou, China \\
lijy727@mail2.sysu.edu.cn} \\

\IEEEauthorblockN{6\textsuperscript{th} Shuai Zhao\IEEEauthorrefmark{1}}
\IEEEauthorblockA{\textit{School of Computer Science and Engineering} \\
\textit{Sun Yat-Sen University}\\
GuangZhou, China \\
zhaosh56@mail.sysu.edu.cn}
\thanks{\IEEEauthorrefmark{1}Corresponding author: Shuai Zhao.}
}

\maketitle

\input{contents/1.abstract}
\input{contents/2.introduction}
\input{contents/3.limitation}
\input{contents/4.framework}

\input{contents/5.rationale}
\input{contents/6.experiment}
\input{contents/7.conclusion}

\balance

\bibliographystyle{IEEEtran}
\bibliography{ref}

\end{document}

%% file: contents/1.abstract.tex
\begin{abstract}

The Large Language Models (LLM) are increasingly being deployed in robotics to generate robot control programs for specific user tasks, enabling embodied intelligence.
Existing methods primarily focus on LLM training and prompt design that utilize LLMs to generate executable programs directly from user tasks in natural language.
However, due to the inconsistency of the LLMs and the high complexity of the tasks, such best-effort approaches often lead to tremendous programming errors in the generated code, which significantly undermines the effectiveness especially when the light-weight LLMs are applied.
This paper introduces a natural-robotic language translation framework that (i) provides correctness verification for generated control programs and (ii) enhances the performance of LLMs in program generation via feedback-based fine-tuning for the programs.
To achieve this, a Robot Skill Language (\lname{}) is proposed to abstract away from the intricate details of the control programs, bridging the natural language tasks with the underlying robot skills.
Then, the \lname{} compiler and debugger are constructed to verify \lname{} programs generated by the LLM and provide error feedback to the LLM for refining the outputs until being verified by the compiler.
This provides correctness guarantees for the LLM-generated programs before being offloaded to the robots for execution, significantly enhancing the effectiveness of LLM-powered robotic applications. 
Experiments demonstrate \fname{} outperforms the existing method under a range of LLMs and tasks, and achieves a high success rate for light-weight LLMs.

\end{abstract}

%% file: contents/2.introduction.tex
\section{Introduction}
\label{sec: introduction}

With the growing demand for intelligent and autonomous robots in a wide range of applications (\eg, smart factories~\cite{harapanahalli2019autonomous}, healthcare~\cite{kyrarini2021survey}, and householding~\cite{sahin2007household}), conventional robotic systems face significant challenges, which can only complete specific tasks in predefined scenarios. To bridge this gap, LLMs (\eg, OpenAI's GPT~\cite{GPT}, Meta's Llama~\cite{Llama}, and Google's Gemma~\cite{team2024gemma}) are deployed in robotics, leveraging their semantic comprehension and contextual reasoning to generate robotic control programs that fulfill the given tasks. 

\begin{figure}[!t]
    \centering
    \includegraphics[width=\linewidth]{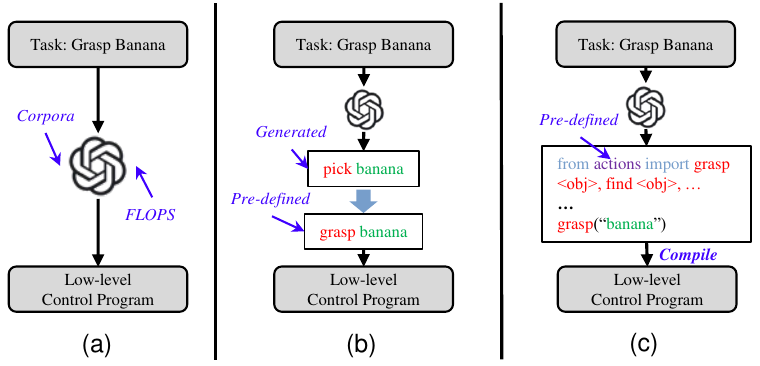}
    \caption{Three paradigms of LLM-powered control program generation: (a) LLMs fine-tuned for low-level control program generation, (b) Task decomposition by LLMs into sub-tasks mapped to the pre-defined robot actions for low-level control programs, and (c) LLMs generate high-level programs compiled into the low-level control programs.}
    \label{fig: system}
    \vspace{-10pt}
\end{figure}

Most of the existing LLM-powered control program generation methods for robotics can be broadly categorized into the following three fundamental paradigms~\cite{brohan2022rt, rana2023sayplan, liang2023code}, as shown in Fig.~\ref{fig: system}.
In Fig.~\ref{fig: system}(a), the pre-trained LLMs are fine-tuned as an end-to-end solution that directly generates low-level control programs for underlying hardware (\eg, motor, sensor) of a robot given specific user tasks.
This approach~\cite{brohan2022rt, driess2023palm} simplifies the workflow of the generation process; however, it requires both enormous datasets and computational resources to tune the LLMs, which are often unavailable in resource-constrained scenarios. To address this, prompt engineering is proposed that utilizes LLMs to transform user tasks into robot actions without model training. As shown in Fig.~\ref{fig: system}(b), the pre-trained LLMs with tailor-designed prompts are applied to decompose user tasks into a series of pre-defined sub-tasks (actions), which are mapped to specific low-level hardware control programs for execution. In Fig.~\ref{fig: system}(c), the LLMs are applied to generate the high-level programs (\eg, in Python) for robotics, which are interpreted into the low-level programs for controlling the hardware.

As discussed above, the LLM fine-tuning approach suffers from notable limitations in applicability due to the need for massive computational resources~\cite{yang2024transfer}. Considering resource-constrained scenarios, prompt engineering for control program generation is preferred due to the elimination of model training. However, with this approach, the generated control program is often prone to programming errors due to the inherent inconsistency of the LLMs and the high complexity of user tasks~\cite{vemprala2023chatgpt}. In addition, the complexity of tasks highlights the challenge of deploying light-weight LLMs on resource-constrained devices, as their performance often falls short of the required standards.

\textbf{Contributions.} To tackle these issues, this paper proposes \fname{}, an LLM-powered natural-to-robotic language translation framework that translates user tasks into executable robot control programs with correctness guarantees. To achieve this, we first develop a high-level Robot Skill Language (\lname{}) that reduces the complexity of program generation for LLMs, effectively bridging user tasks with robot skills by removing the robot control program details. This allows the LLM to directly generate \lname{} program without considering details of robot control programs.
To verify the correctness of the generated \lname{} program, an \lname{} compiler is constructed that verifies the correctness for generated \lname{} programs, providing correctness guarantees for robot control programs compiled from \lname{} programs.
In addition, to address the detected programming errors, a \lname{} debugger is developed that provides a feedback-based fine-tuning mechanism to refine the program. This effectively improves the success rate for generating compiler-verified control programs, especially for the light-weight LLMs on resource-constrained robots.

The experimental results demonstrate that \fname{} outperforms existing methods~\cite{singh2023progprompt} by 53.6\% on average in terms of the success rate of control program generation. In particular, \fname{} increases the success rate by 91.6\% without model training by enabling the feedback-based program fine-tuning, which effectively detects and corrects errors in the generated \lname{} program before sending for execution. In addition, experiments demonstrate the effectiveness of \fname{} for light-weight LLMs, attaining a success rate of 92\% with an LLM of 2B parameters~\cite{team2024gemma}.

%% file: contents/3.limitation.tex
\section{The State-of-art and Limitation}
\label{sec: limitation}

\begin{figure*}[!htbp]
	\centering
	\includegraphics[width=\textwidth]{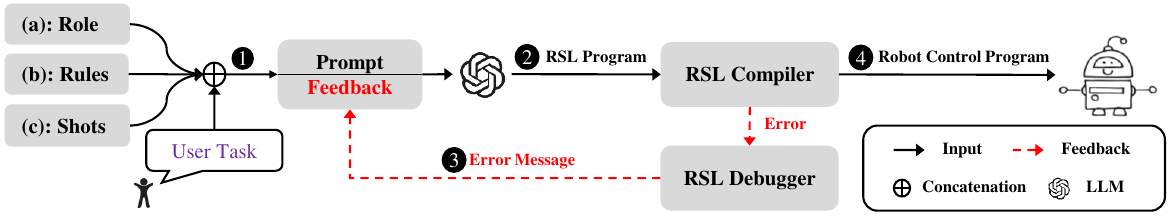}
	\caption{The overall structure and workflow of \fname{} framework.}
	\label{fig: framework}
    \vspace{-10pt}
\end{figure*}

As trained on extensive corpora, pre-trained LLMs demonstrate remarkable capabilities in code generation for high-level languages (\eg, C, Python). To accommodate LLMs to generate low-level control programs in robotic applications, LLMs, which are fine-tuned with datasets collected from specific robots and scenarios, have been emerging, including Google's RT-1~\cite{brohan2022rt}, Google's PaLM-E~\cite{driess2023palm}, and BIGAI's LEO~\cite{huang2023embodied}. The LLMs process user input, and directly generate low-level control programs to complete user tasks. However, this suffers from the following limitation due to the constrained resources.

\vspace{-.4em}
\begin{limitation}{Resource Constraint}{limitation-1}
\vspace{-.4em}
The fine-tuning for LLMs~\cite{brohan2022rt,driess2023palm,huang2023embodied} lacks practicality due to their reliance on scenario-specific corpora and extensive computational resources~\cite{yang2024transfer}.
\vspace{-.4em}
\end{limitation}
\vspace{-.4em}

In order to reduce the reliance on datasets and computation resources for applicability, a portion of existing work utilizes prompt engineering to enable LLMs in robotic applications. The work~\cite{huang2022language} validates that well-designed prompts enable LLMs to decompose high-level tasks into sub-tasks executable by robots. Based on this, existing methods~\cite{doasican, huang2022language, hazra2024saycanpay, huang2023embodied, huang2022inner, song2023llm} decompose user tasks into sub-tasks that are mapped to pre-defined robot actions for low-level control program generation. For instance, Language Planner~\cite{huang2022language} generates mid-level action plans in natural language, mapping them to pre-defined actions. This leads to the following limitation.

\vspace{-.4em}
\begin{limitation}{Capability Requisition}{limitation-2}
\vspace{-.4em}
Task decomposition~\cite{hazra2024saycanpay, huang2022inner, song2023llm} requires superior capabilities of LLMs to generate reasonable and coherent sub-tasks, making it challenging for the light-weight LLMs (\eg, Gemma2-2b and Gemma2-9b)~\cite{huang2022language}.
\vspace{-.4em}
\end{limitation}
\vspace{-.4em}

As discussed above, LLMs excel at code generation for high-level languages, which is leveraged by some existing work~\cite{liang2023code, honerkamp2024language, singh2023progprompt, rana2023sayplan, lin2023text2motion, tanneberg2024help, wang2024llm, yang2023octopus}. The above methods utilize LLMs to receive user tasks and generate high-level language programs, which are then compiled into low-level robot control programs. ProgPrompt~\cite{singh2023progprompt} uses a Pythonic representation to enhance the program generation of LLMs. This method relies on high-level language generation through LLMs and leads to the following limitation.

\vspace{-.4em}
\begin{limitation}{LLMs Inconsistency}{limitation-3}
\vspace{-.4em}
High-level language generation~\cite{singh2023progprompt, schafer2023empirical, xu2024lecprompt} are prone to programming errors due to the inconsistency of LLMs, significantly limiting the effectiveness due to the lack of correctness guarantees~\cite{vemprala2023chatgpt}.
\vspace{-.4em}
\end{limitation}
\vspace{-.4em}

In addition, most existing methods are tailored for specific robots~\cite{brohan2022rt} and scenarios (\eg, simulating household VirtualHome~\cite{puig2018virtualhome}), or depend on LLMs with superior capabilities, while struggling with the inconsistency of LLMs~\cite{vemprala2023chatgpt}. These limitations make them challenging to apply in real-world robotic applications, in which light-weight LLMs on resource-constrained robots generate control programs that face the inconsistency of LLMs. To address this, we propose, \fname{}, an LLM-powered natural-to-robotic language translation framework that offers correctness guarantees for programs and improves the success rate for light-weight LLMs.

%% file: contents/4.framework.tex
\section{Framework Overview and Requirements}
\label{sec: framework}

To clarify how the \fname{} works, we provide an overview that translates user tasks to robot control programs, shown as four stages in Fig.~\ref{fig: framework}. As the first work that provides correctness guarantees for robot control program generation, we focus on the design and construction of \fname{} while considering the following preliminaries and assumptions. 
This provides the groundwork for future extensions that include environmental interactions, real-world deployment, multi-robot collaboration, etc. Below, we present the requirements for \fname{} and discuss the solutions in Sec.~\ref{sec: rationale}.

\subsection{Preliminaries and Assumptions}
\label{sec: assumption}

To facilitate the construction, we establish the following preliminaries and assumptions for the \fname{}:
\begin{enumerate}[label=P$\arabic*$., ref=$\arabic*$]
    \item \label{preliminary1} We adopt a uniform prompt format sourced from OpenAI's API and extend to a range of LLMs with existing inference tools (\eg, llama.cpp~\cite{llama-cpp}).
    \item \label{preliminary2} Following prior work~\cite{pages2016tiago}, we assume that robots are built on the Robot Operating System (ROS)~\cite{Quigley09} and equipped with Python-based interfaces for robot control programming. These Python interfaces are encapsulated as discrete functions without self-decision-making.
    \item \label{preliminary3} Advanced language mechanisms (\eg, conditional and loop statements) and robotic capabilities (\eg, dynamic environment monitoring) are deferred to future work, allowing us to prioritize the design and validation of the proposed translation framework for most tasks, thus providing a proof-of-concept for compiler-verified language translation by LLMs.
\end{enumerate}

\subsection{Framework Overview}
\label{sec: overview}

For robot control program generation, resolving the inconsistency of LLMs is critical to guarantee consistent outputs for underlying robots. To address this, we propose \fname{}, an LLM-powered framework for natural-to-robotic language translation with correctness guarantees. Considering the deployment of the \fname{} framework, we propose a feedback-based fine-tuning approach to minimize the requisitions for superior capabilities of LLMs. This method can achieve a high success rate when light-weight LLMs are applied. As illustrated in Fig.~\ref{fig: framework}, \fname{} employs a structured four-stage phase to establish an entire workflow.

\textit{Stage \protect\blackcircled{1}: Prompt Construction and \lname{} Generation.} In our work, the prompt is initialized with the system message, shots (optional), and user task. The system message $\mathcal{M}$ defines the role of LLMs, expected generation specifications, and target language rules to the LLM, effectively guiding the generation of \lname{} programs utilized in Stage \protect\blackcircled{2}, as shown in Fig.~\ref{fig: framework}. The shots $\mathcal{S}$ provide output templates comprising selected user tasks and their corresponding \lname{} programs, constraining the LLM to generate outputs consistent with the desired format. Given a user task $\tau$ expressed in natural language, the prompt $\mathcal{P}$ is constructed by directly concatenating the system message, shots, and user task in sequence. The resulting prompt $\mathcal{P}$ serves as the query to the LLM, which generates the \lname{} program $\chi$ for the user task $\tau$ without human intervention.

\textit{Stage \protect\blackcircled{2}: \lname{} Compilation and Validation.} The compiler typically receives a high-level language program as input and then compiles it into a lower-level language, detecting programming errors if the program violates the language rules. For the generated \lname{} program, the \lname{} compiler translates the \lname{} program $\chi$ into a robot control program $\mathcal{R}$ and identifies hidden errors $\mathcal{E}$ within the program. The compiler ensures the correctness of the \lname{} program and compiles it into an executable program for the underlying robot in Stage \protect\blackcircled{4}, thereby providing compiler-verified correctness guarantees for the generated robot control programs. However, if the \lname{} compiler detects errors $\mathcal{E}$, the \fname{} framework enters a feedback-based fine-tuning iteration for \lname{} program debugging in Stage \protect\blackcircled{3}.

\textit{Stage \protect\blackcircled{3}: Feedback Composition and \lname{} Fine-Tuning.} The debugger is critical in identifying and resolving errors within \lname{} programs, improving the success rate of program generation without model training. If errors $\mathcal{E}$ are detected in Stage \protect\blackcircled{2}, the error messages (e.g., a missing semicolon) generated by the \lname{} debugger are incorporated into the prompt $\mathcal{P}$ to form the feedback $\mathcal{F}$ for LLM re-evaluation. As shown in Fig.~\ref{fig: framework}, the feedback $\mathcal{F}$ is utilized by LLMs to fine-tune the \lname{} program $\chi$ for solving programming errors. In our work, Stage \protect\blackcircled{2} and Stage \protect\blackcircled{3} operate in a closed fine-tuning loop to improve the success rate of program generation.

\textit{Stage \protect\blackcircled{4}: Robot Control Program Execution.} If the \lname{} program $\chi$ successfully passes correctness verification in Stage \protect\blackcircled{2} or exits the fine-tuning iteration between Stage \protect\blackcircled{2} and Stage \protect\blackcircled{3}, the compiled robot control program is deployed to the underlying robot for task completion. The control program is automatically translated into the low-level control program to control robot hardware (\eg, motor, sensor), enabling the robot to perform the corresponding actions.

\subsection{Requirements for \fname{}}
\label{sec: requirement}

Compared with \fname{}, existing work utilizes formatted error messages with intricate function invocation from compilers aimed to fix high-level language errors~\cite{schafer2023empirical, xu2024lecprompt} and employed feedback for robot planning to complete tasks~\cite{joublin2024copal}, facing runtime risks during execution. To the best of our knowledge, our work is the first to achieve a high success rate of task completion with light-weight LLMs, addressing LLM inconsistencies and providing correctness guarantees for practical applications. In our work, we independently designed and implemented the \fname{} framework. The requirements for building \fname{} are summarized as follows:

\begin{enumerate}[label=R$\arabic*$., ref=\arabic*]
    \item \label{requirement1} To decompose user tasks into robot-executable skills, \lname{} must represent the capabilities of robots, offering intuitive and actionable representations for LLMs.  
    \item \label{requirement2} To enable feedback-based fine-tuning with acceptable response times, the \lname{} debugger should distill program errors into concise, intuitive messages, omitting unnecessary details and providing clear guidance for LLMs.  
    \item \label{requirement3} To support diverse robots and scenarios, \fname{} must quickly adapt to new capabilities and environments, ensuring compatibility and scalability of \fname{}.
\end{enumerate}

%% file: contents/5.rationale.tex
\begin{table*}[!htbp]
    \caption{\lname{} Lexical and Syntax Design}
    \label{tab: language}
    \centering
    \resizebox{\textwidth}{!}{
        \begin{tabular}{ll
        >{\columncolor[HTML]{d8ffd8}}l 
        >{\columncolor[HTML]{ffd8d8}}l }
            \hline
            \multicolumn{1}{c}{\cellcolor[HTML]{C0C0C0}\textbf{Robot Capability}} & \multicolumn{1}{c}{\cellcolor[HTML]{C0C0C0}\textbf{Description}}    & \multicolumn{1}{c}{\cellcolor[HTML]{C0C0C0}\textbf{Keyword}} & \multicolumn{1}{c}{\cellcolor[HTML]{C0C0C0}\textbf{Syntax}} \\ \hline
            \textbf{Forward}                                                      & Moves the robot forward by a specified distance.                    & FORWARD                                                      & FORWARD NUMBER;                                             \\ \hline
            \textbf{Backward}                                                     & Moves the robot backward by a specified distance.                   & BACKWARD                                                     & BACKWARD NUMBER;                                            \\ \hline
            \textbf{Turn Left}                                                    & Rotates the robot left by a specified angle.                        & TURNLEFT                                                     & TURNLEFT NUMBER;                                            \\ \hline
            \textbf{Turn Right}                                                   & Rotates the robot right by a specified angle.                       & TURNRIGHT                                                    & TURNRIGHT NUMBER;                                           \\ \hline
            \textbf{Look Up}                                                      & Tilts the camera or sensor upward by a specified angle.             & LOOKUP                                                       & LOOKUP NUMBER;                                              \\ \hline
            \textbf{Look Down}                                                    & Tilts the camera or sensor downward by a specified angle.           & LOOKDOWN                                                     & LOOKDOWN NUMBER;                                            \\ \hline
            \textbf{Look Left}                                                    & Rotates the camera or sensor left by a specified angle.             & LOOKLEFT                                                     & LOOKLEFT NUMBER;                                            \\ \hline
            \textbf{Look Right}                                                   & Rotates the camera or sensor right by a specified angle.            & LOOKRIGHT                                                    & LOOKRIGHT NUMBER;                                           \\ \hline
            \textbf{Perceive}                                                     & Builds a SLAM map of the environment.                               & PERCEIVE                                                     & PERCEIVE;                                                   \\ \hline
            \textbf{Approach}                                                     & Identifies and navigates to a specified object.                     & APPROACH                                                     & APPROACH OBJECT;                                            \\ \hline
            \textbf{GoTo}                                                         & Navigates to a given coordinate.                                    & GOTO                                                         & GOTO NUMBER, NUMBER;                                        \\ \hline
            \textbf{Grasp}                                                        & Picks up a specified object with the robotic arm.                   & GRASP                                                        & GRASP OBJECT;                                               \\ \hline
            \textbf{Others}                                                       & Includes numbers, objects, and delimiters following C syntax rules. & -                                                            & -                                                           \\ \hline
        \end{tabular}
    }
    \vspace{-10pt}
\end{table*}

\begin{figure}[!htbp]
	\centering
	\includegraphics[width=\linewidth]{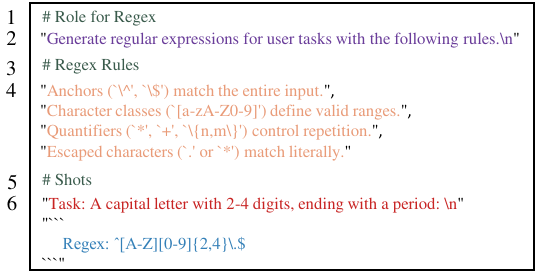}
	\caption{A prompt for Regex Generation.}
	\label{fig: regex}
    \vspace{-10pt}
\end{figure}

\section{Rationale and Detail for \fname{}}
\label{sec: rationale}

As discussed in Sec.~\ref{sec: requirement}, the design and implementation of the \fname{} need to satisfy requirements that affect the effectiveness and efficiency of translation from user tasks to robot control programs. In this section, we present design rationales and implementation details about each component of \fname{} to illustrate how to meet the requirements, providing thorough insights to support further research and development. In addition, characteristics of \fname{} are discussed to address limitations mentioned in Sec.~\ref{sec: limitation}.


\subsection{Prompt Setting}
\label{sec: prompt}

As described in Sec.~\ref{sec: overview}, prompts are constructed with the system message and (optional) shots, which provide task-related foundational information and generation templates for LLMs. To demonstrate the system message and shots, we use \textbf{Regular Expression} as a concrete example to illustrate how LLMs can translate user tasks in Fig.~\ref{fig: regex}.

\begin{figure}[!htbp]
	\centering
	\includegraphics[width=\linewidth]{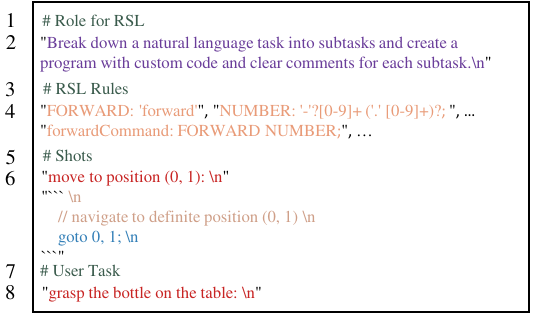}
	\caption{An illustrative example of the input prompt ((a) $\oplus$ (b) and optional (c) in Fig.~\ref{fig: framework}).}
	\label{fig: prompt}
    \vspace{-10pt}
\end{figure}

As demonstrated above, shots provide representative examples to enhance the performance of LLMs for subsequent user tasks. In our work, one shot is provided per \lname{} keyword; however, shots are not mandated in \fname{}, as error feedback can fine-tune generated programs. Experiments with zero-shot prompts in Sec.~\ref{sec: zero-shot} validate this. As shown in Fig.~\ref{fig: prompt}, the system message, the optional shots, and the task are concatenated to form the prompt for LLMs.

\subsection{\lname{} Design and Compilation}
\label{sec: compiler}

The \lname{} language with its compiler forms the foundation for translation from user tasks to robot control programs with correctness guarantees in \fname{}. In this section, we illustrate the rationales of \lname{} design and how to implement \lname{} compiler. Guided by R\ref{requirement1}, \lname{} abstracts robot capabilities to achieve translation from \lname{} programs to robot control programs, while \lname{} compiler provides correctness guarantees within \lname{} program compilation.

\textit{\lname{}.} The keywords of \lname{} are derived from robot capabilities, where each keyword is associated with one or more interfaces. These keywords abstract away from the details of robot control programs, highlighting their semantics of robot skills for LLMs to facilitate \lname{} program generation. As shown in Tab.~\ref{tab: language}, the keyword meaning is intuitive, designed to express the robot's functionalities in natural language. The additional elements in \lname{}, such as the identifiers, numbers, and comment style, follow the conventions in C programming.

\vspace{-.4em}
\begin{characteristic}{\lname{} Simplicity}{characteristic-1}
\vspace{-.4em}
\lname{} encapsulates the intuitive semantics of robot skills, enabling LLMs to generate \lname{} programs for user tasks without model fine-tuning, addressing Limitation 1.
\vspace{-.4em}
\end{characteristic}
\vspace{-.4em}

The syntax of \lname{} is command-based, where each \lname{} statement consists of a keyword and optional parameters, and ends with a semicolon. Tab.~\ref{tab: language} outlines the syntax rules of \lname{}, showing the usage and parameters for each keyword. We enforce that the numeric parameters must be positive to ensure correct robot actions (\eg, the forward instruction would not cause the robot to move backward). Identifiers represent objects manipulated by the robot with actions represented by specific keywords.

In addition, the conciseness of the \lname{} reduces the generation length for LLMs compared with conventional high-level language, which is advantageous for resource-constrained devices where token generation may be limited to one per second. However, \lname{} programs generated by LLMs are mostly accompanied by illustrative content. Considering the scope of this paper, the solution for \lname{} program generation and detailed analysis are deferred to future work.

\begin{figure}[!htbp]
	\centering
	\includegraphics[width=\linewidth]{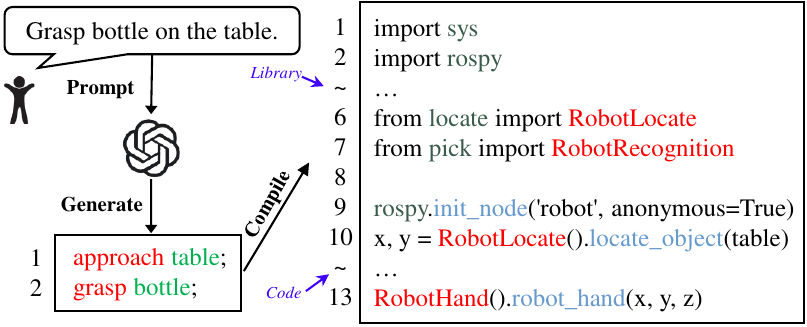}
	\caption{{A translation from a user task to a robotic program.}}
	\label{fig: translation}
    \vspace{-5pt}
\end{figure}

\textbf{\lname{} Compiler.}
The \lname{} compiler verifies the syntax correctness of \lname{} programs, binds \lname{} rules to robot control interfaces, and generates executable robot control programs. In our work, lexical and syntactic rules of \lname{} are defined in regular expressions (see Tab.~\ref{tab: language}), which are compatible with existing tools (\eg, ANTLR~\cite{parr1995antlr}) that automatically generate the lexer and parser for \lname{}. The generated lexer and parser analyze \lname{} programs that are separated as individual tokens and parsed as an abstract syntax tree (AST) using LL(1) syntax analysis. A custom code generator, implemented with only a few lines of code (LOC), processes the AST in a depth-first manner, translating sub-trees (\ie, \lname{} statements) into robot control programs. The automatic generation for lexer and parser, and the compact code generator support R\ref{requirement3}.

As illustrated in Fig.~\ref{fig: translation}, the LLM generates the corresponding \lname{} program given a user task. The \lname{} program serves as input for \lname{} compiler, then is compiled to an executable robot control program. In our work, \lname{} compiler compiles \lname{} program into a Python program, automatically importing necessary libraries and sequentially invoking required functions to complete tasks.

\vspace{-.4em}
\begin{characteristic}{Correctness Guarantees}{characteristic-2}
\vspace{-.4em}
The \lname{} compiler provides correctness verification for generated \lname{} programs to solve the inconsistency of LLMs, effectively addressing Limitation 3.
\vspace{-.4em}
\end{characteristic}
\vspace{-.4em}

\subsection{\lname{} Debugger}
\label{sec: debugger}

During the \lname{} compilation, programming errors are detected by \lname{} compiler for \lname{} programs, constructing associated error messages for LLMs by \lname{} debugger. Error messages enable LLMs to debug \lname{} programs, improving the success rate for \lname{} program generation.

\begin{figure}[!htbp]
	\centering
	\includegraphics[width=\linewidth]{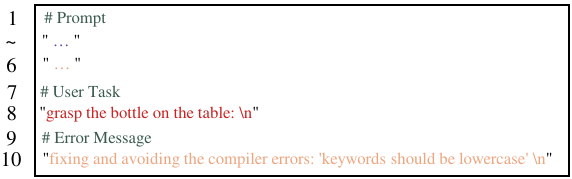}
	\caption{{An example of the feedback for fine-tuning.}}
	\label{fig: feedback}
    \vspace{-10pt}
\end{figure}

\textbf{Feedback-based Tuning.} Based on the error messages, we propose a feedback-based fine-tuning method that iteratively refines generated \lname{} programs. If errors are spotted by \lname{} compiler, error messages will be produced by \lname{} debugger. As shown in Fig.~\ref{fig: feedback}, messages are concatenated with prompts to form error feedback for LLMs, fixing existing errors through regeneration of the \lname{} programs. The fine-tuning iteration finishes until the program passes the compiler verification. As illustrated in Sec.~\ref{sec: zero-shot}, the feedback-based tuning effectively addresses the lack of semantic information for ambiguous tasks and zero-shot scenarios, improving the applicability of the \fname{} across different scenarios.

\vspace{-.4em}
\begin{characteristic}{Capacity Enhancement}{characteristic-3}
\vspace{-.4em}
\fname{} employs a feedback-based fine-tuning method that enables light-weight LLMs to achieve a high success rate on resource-constrained devices, addressing Limitation 2.
\vspace{-.4em}
\end{characteristic}
\vspace{-.4em}

\textbf{Error Messages.} Traditional error messages produced by compilers often contain sophisticated technical details, preventing an untrained LLM from understanding and fixing the errors. To overcome this, we design semantic-intuitive error messages based on the \lname{} lexical and syntactic rules, with the line number and exact token highlighted. The lexical error messages are classified into five categories for every lexeme in \lname{}, including the keyword, identifier, number, character, and comment. Four types of syntactic error messages are constructed for syntactic errors in \lname{}. These error messages are presented in natural language with intuitive semantics for LLMs, as demonstrated in Tab.~\ref{tab: message}. Based on the above intuitive error messages, \fname{} improves the success rate with feedback-based fine-tuning, satisfying the R\ref{requirement2}.

\begin{table}[!htbp]
    \caption{Error Message Design for \lname{}}
    \label{tab: message}
    \centering
    \resizebox{\linewidth}{!}{
        \begin{tabular}{l
        >{\columncolor[HTML]{ffd8d8}}ll }
            \hline
            \multicolumn{1}{c}{\cellcolor[HTML]{C0C0C0}\textbf{Type}} & \multicolumn{1}{c}{\cellcolor[HTML]{C0C0C0}\textbf{Error Message}} & \multicolumn{1}{c}{\cellcolor[HTML]{C0C0C0}\textbf{Example}} \\ \hline
            \textbf{Keyword}                                          & Keywords should be lowercase.                                      & APPROACH table;                                              \\ \hline
            \textbf{Identifier}                                       & The identifier is illegal.                                         & approach 3apple;                                             \\ \hline
            \textbf{Number}                                           & The number is illegal.                                             & forward 123.23.45;                                           \\ \hline
            \textbf{Character}                                        & The \$ is an illegal character.                                    & \$forward 1;                                                 \\ \hline
            \textbf{Comment}                                          & This comment has errors.                                           & / This is a comment                                          \\ \hline
            \textbf{Command}                                          & The command (keyword) is illegal.                                  & move 1.5;                                                    \\ \hline
            \textbf{Parameter}                                        & Parameter types of the command are invalid.                        & forward table;                                               \\ \hline
            \textbf{Quantity}                                         & The number of parameters is illegal.                               & goto 1;                                                      \\ \hline
            \textbf{Semicolon}                                        & The statement must end with a semicolon.                           & approach table                                               \\ \hline                                       
        \end{tabular}
    }
\end{table}

\subsection{Robot Capsulation}
\label{sec: robot}

In \fname{}, any requirements for decision-making or higher-level reasoning are delegated to the LLM, making robots receive definite parameters to execute specific actions. Based on definite robot skills, lexical and syntax rules can be easily constructed as keywords representing robot actions, with required parameters. As discussed above, rules of \lname{} in regular expressions and the code generator are implemented within hundreds of LOC, making \fname{} easily customized or extended to accommodate various robotic platforms.

%% file: contents/6.experiment.tex
\section{Evaluation}
\label{sec: evaluation}

To evaluate \fname{}, we compare it with ProgPrompt~\cite{singh2023progprompt}, which utilizes a Pythonic prompt for LLMs to generate robot control programs, in terms of the success rate, and accuracy of generated programs for \fname{}. Current task decomposition methods are not included due to the lack of open-source implementations and the incompatibility with the evaluation objectives for \fname{}~\cite{doasican, hazra2024saycanpay}.

\begin{table}[!htbp]
    \caption{Examples of User Tasks for Evaluation.}
    \label{tab: task}
    \centering
    \renewcommand{\arraystretch}{1.2}
    \resizebox{\columnwidth}{!}{%
        \begin{tabular}{p{0.15\columnwidth}|p{0.75\columnwidth}}
        \hline
        \multicolumn{1}{c|}{\cellcolor[HTML]{C0C0C0}Group} & \multicolumn{1}{c}{\cellcolor[HTML]{C0C0C0}Task} \\ \hline
        \multirow{6}{*}{\textbf{{\renewcommand{\arraystretch}{1} \begin{tabular}[c]{@{}c@{}} ~~Simple~~\\~Tasks\end{tabular}}}} 
        & Turn left 1 rad.  \\
        & Turn right 3 rads.  \\
        & Approach the door.  \\
        & Move forward 5 meters.  \\
        & Move backward 3 meters.  \\
        & Turn around a half circle. \\ \hline
        \multirow{4}{*}{\textbf{{\renewcommand{\arraystretch}{1} \begin{tabular}[c]{@{}c@{}} \hspace{-0.4em}Ambiguous\\Tasks~\end{tabular}}}} 
        & Go forward a little.  \\
        & Go forward further.  \\
        & Go forward a long distance.  \\
        & Turn around several circles.  \\ \hline
        \multirow{7}{*}{\textbf{{\renewcommand{\arraystretch}{1} \begin{tabular}[c]{@{}c@{}} Multi-step\\Tasks\end{tabular}}}} 
        & Move forward 10 meters, then look around. \\
        & Approach the workbench, then grasp a tool. \\
        & Grasp the book, then move backward 2 meters. \\
        & Move forward 5 meters, perceive, then turn right. \\
        & Forward 2 meters, turn right 2 rads, and forward 1 meter. \\
        & Move forward 2 meters, grasp the banana, turn left, then move forward 3 meters. \\ \hline
        \multirow{12}{*}{\textbf{{\renewcommand{\arraystretch}{1} \begin{tabular}[c]{@{}c@{}} ~Complex\\~Tasks\end{tabular}}}} 
        & Grasp the bottle on the table and go back for me. \\
        & Forward 4 meters, turn right, perceive environment. \\
        & Repeat forward 1 meter and turn right 1.57 rads 4 times. \\
        & Perceive the environment, approach the ball, and grasp it. \\
        & Approach the table, look for a cup, and grasp it carefully. \\
        & Approach the table, return to the origin in the coordinate. \\
        & Move forward 3 meters, turn left, then move another 2 meters, then turn right. \\
        & Repeat moving forward 1 meter and turning right 3 times for a triangular trajectory. \\
        & Move forward 2 meters, turn right, look around; repeat this pattern until a full circle. \\ \hline
    \end{tabular}
    }
    \vspace{-10pt}
\end{table}

\subsection{Experimental Setup and Metrics}
\label{sec: setup}

\textbf{Setup.} 
The \fname{} is deployed on an Intel i5-13400 CPU with an Nvidia RTX 4060Ti GPU. The Tiago~\cite{pages2016tiago} robot is applied for evaluation, equipped with ROS 1.0 and Python interfaces. The competing methods are evaluated under twenty-five user tasks, which are designed based on the skills of the Tiago robot. To assess the performance of \fname{} under different LLMs, Gemma2-2b~\cite{team2024gemma}, Gemma2-9b, Llama-70b~\cite{Llama}, Gemini-1.5-Flash~\cite{team2023gemini}, and GPT-4o~\cite{achiam2023gpt} are considered in the evaluation, covering a large scale of parameter size from 2 billion to 1.76 trillion (\ie, GPT-4o)~\cite{enwiki:1246495302}. 

\textbf{Dataset.} 
Experimental tasks are categorized into four groups: (i) simple tasks, (ii) ambiguous tasks, (iii) multi-step tasks, and (iv) complex tasks, as shown in Tab.~\ref{tab: task}. The tasks are designed with varying levels of complexity, demanding advanced capabilities from LLMs while addressing inconsistencies in their outputs. Using this task set, we aim to validate the effectiveness of \fname{}.

\textbf{Metric.} 
To quantify the performance of the \fname{} on the above tasks, three major metrics of the generated programs are considered: (i) \textbf{success rate}, verifying whether generated programs pass their compiler and conform to a uniform format that can be automatically executed; (ii) \textbf{accuracy}, assessing whether the program completes the requested task without the consideration of robot location and pose; and (iii) \textbf{pass}, representing the average number of iterations required for the LLM to generate the correct program, when feedback-based fine-tuning is enabled.

\subsection{Overall Performance Comparison under Varied LLMs}
\label{sec: comparison}

\begin{figure}[!htbp]
\centering
    \includegraphics[width=\columnwidth]{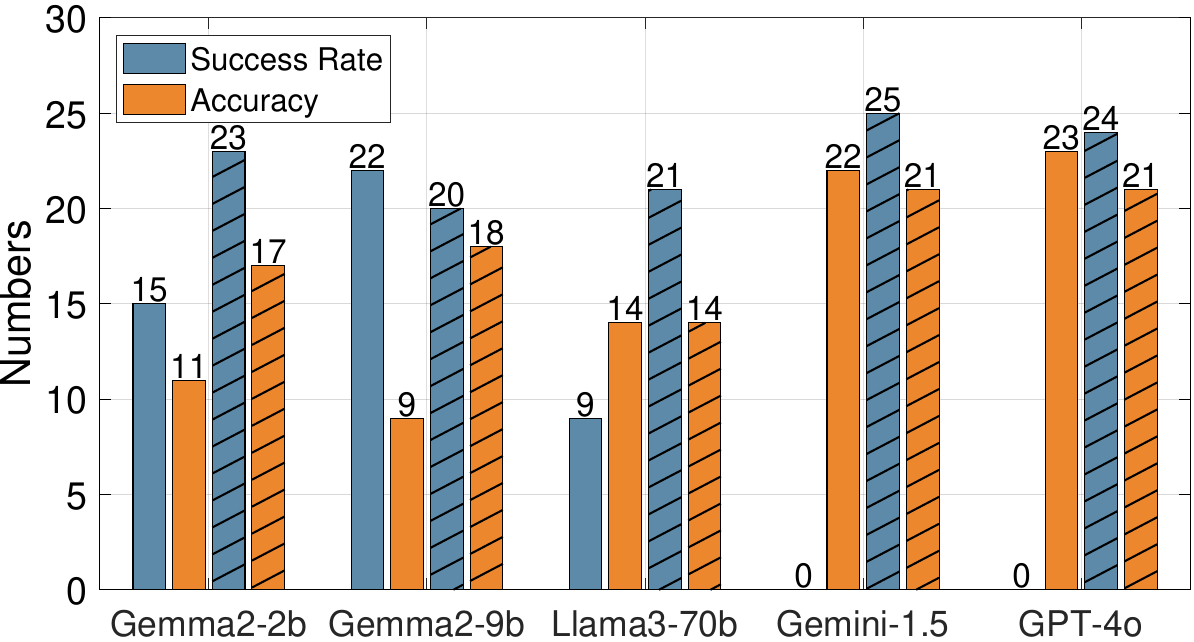}
    \caption{Success Rate and Accuracy of Methods \textit{(Bars without markers -- ProgPrompt; Bars with markers - \fname{};)}.}
    \label{fig: llms}
    \vspace{-10pt}
\end{figure}

The \fname{} is evaluated against the ProgPrompt~\cite{singh2023progprompt}, which constructs a Python-structured prompt to import available objects and Tiago's interfaces for control program generation. The ProgPrompt can integrate with different LLMs, allowing us to evaluate and compare the effectiveness of both methods across a wide range of LLMs. Based on experimental tasks, Fig.~\ref{fig: llms} reports the number of executable programs and accurate programs generated by both methods under the considered LLMs, respectively.

As shown by the results, we first observe that the \fname{} constantly outperforms ProgPrompt, achieving an improvement of 53.6\% in success rate and 9.6\% in accuracy on the average case. Notably, \fname{} exceeded ProgPrompt by 30\% in accuracy under two light-weight LLMs (\ie, Gemma2-2b and Gemma2-9b). This observation demonstrates that the \fname{} effectively facilitates program generation of LLMs by abstracting away from the complex details of robot control programs, validating Characteristic 1. Interestingly, the ProgPrompt with the two largest and most powerful LLMs (\ie, Gemini-1.5-Flash and GPT-4o) showed the lowest success rate but with high accuracy. This means that without effective constraints and guidance, such models often produce control programs in various forms (e.g., classes instead of functions), leading to execution failures without additional manual adjustments.

As for the \fname{}, it achieves at least 80\% success rate and 56\% accuracy across all LLMs. In general, the success rate of \fname{} is consistently higher than its accuracy for each LLM, translating the understanding of user tasks into correct control programs, and validating Characteristic 2. With Gemini-1.5-Flash and GPT-4o applied, \fname{} can produce correct programs in a single pass for nearly all user tasks. Notably, compared to GPT-4o, the light-weight LLMs (\ie, Gemma2-2b and Gemma2-9b) demonstrate competitive performance on \fname{} by utilizing the feedback-based fine-tuning mechanism (\ie, with more than one pass), validating Characteristic 3. This illustrates the effectiveness of the constructed \lname{} debugger, and more importantly, the applicability of \fname{} on resource-constrained devices.

\subsection{Benefits of Semantic-intuitive Error Messages}
\label{sec: intuitive}

To validate the effectiveness of error messages (See Tab.~\ref{tab: message}), an experiment is conducted to evaluate the semantic-intuitive error messages constructed in Sec.~\ref{sec: debugger}. The \fname{} with the original error messages produced by ANTLR~\cite{parr1995antlr} is applied as the baseline method. The Gemma2-2b is applied for both methods. As shown in Tab.~\ref{tab: error}, \fname{} outperforms the baseline in both success rate and accuracy while requiring fewer passes on average. This indicates that semantic-intuitive error messages can aid the LLM in understanding the type and the location of errors, providing effective guidance for the LLM during the fine-tuning process.

\begin{table}[!htbp]
    \caption{Effectiveness of Customized Error Messages}
    \label{tab: error}
    \centering
    \resizebox{.78\columnwidth}{!}{
    \begin{tabular}{cccc}
        \hline
        \textbf{Method}   & \textbf{Success Rate}   & \textbf{Accuracy}  & \textbf{Pass}    \\ \hline
        Baseline          & 22~/~25                 & 16~/~25            & 1.56             \\ 
        \fname{}          & 23~/~25                 & 17~/~25            & 1.4              \\ \hline
    \end{tabular}}
    \vspace{-10pt}
\end{table}

\subsection{Performance of \fname{} with Zero Shot}
\label{sec: zero-shot}

The prompt often requires modifications to accommodate diverse scenarios. However, it can be challenging for users to provide specific shots that adapt robot control programs into an expected format (\eg, a complete function replacing code statements). To investigate the performance of \fname{} in such situations, an experiment is conducted on \fname{} with zero shot under the GPT-4o, which possesses the necessary ability to comprehend the semantics of user tasks compared to other LLMs such as Gemma and Llama.

As shown in Tab.~\ref{tab: zero-shot}, \fname{} with zero shot shows a slightly lower success rate and a significant drop in accuracy, which is expected due to the absence of detailed semantic information. However, compared to the success rate in one pass (\ie, without feedback-based fine-tuning), \fname{} with zero shot significantly increases the success rate by 91.6\%. This experiment again verifies the effectiveness of the design of \fname{}, and demonstrates its applicability for scenarios where no shots can be provided.

\begin{table}[!htbp]
    \caption{Performance of \fname{} with Zero Shot}
    \label{tab: zero-shot}
    \centering
    \resizebox{\columnwidth}{!}{
    \begin{tabular}{ccccc}
        \hline
        \textbf{Method}                                                         & \textbf{Success Rate}     & {\renewcommand{\arraystretch}{1} \begin{tabular}[c]{@{}c@{}} \textbf{Success Rate} \\ \textbf{(One Pass)} \end{tabular}}    & \textbf{Accuracy}     & \textbf{Pass}     \\ \hline
        {\renewcommand{\arraystretch}{1} \begin{tabular}[c]{@{}c@{}} \fname{} with \\ Zero Shot \end{tabular}}    & 23~/~25                   & 12~/~25                                                                                   & 17~/~25               & 2            \\
        \fname{}                                                                & 24~/~25                   & 24~/~25                                                                                   & 21~/~25               & 1.16              \\ \hline
    \end{tabular}}
    \vspace{-10pt}
\end{table}

\subsection{Effectiveness of Feedback-Based Tuning}
\label{sec: tuning}

\begin{figure}[!htbp]
    \centering
    \subfigure[]{\includegraphics[width=0.24\textwidth]{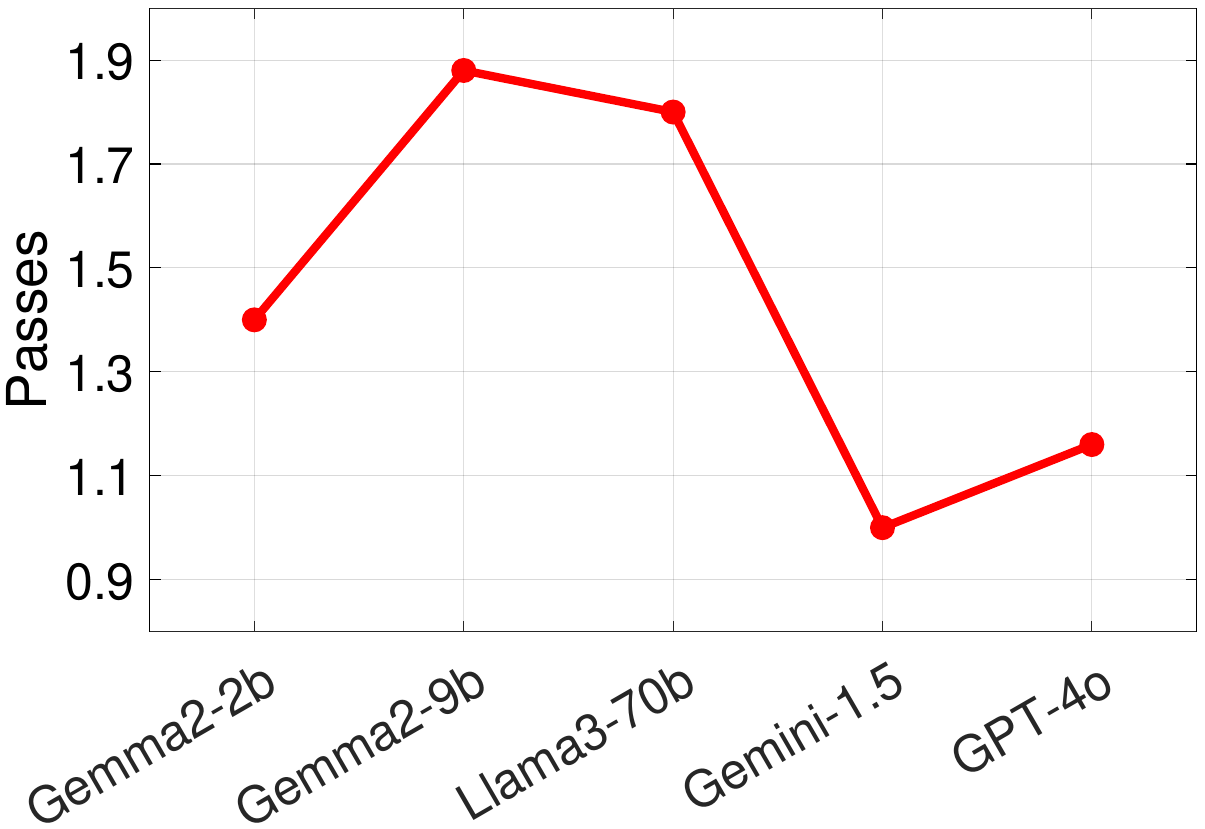}} 
    \subfigure[]{\includegraphics[width=0.24\textwidth]{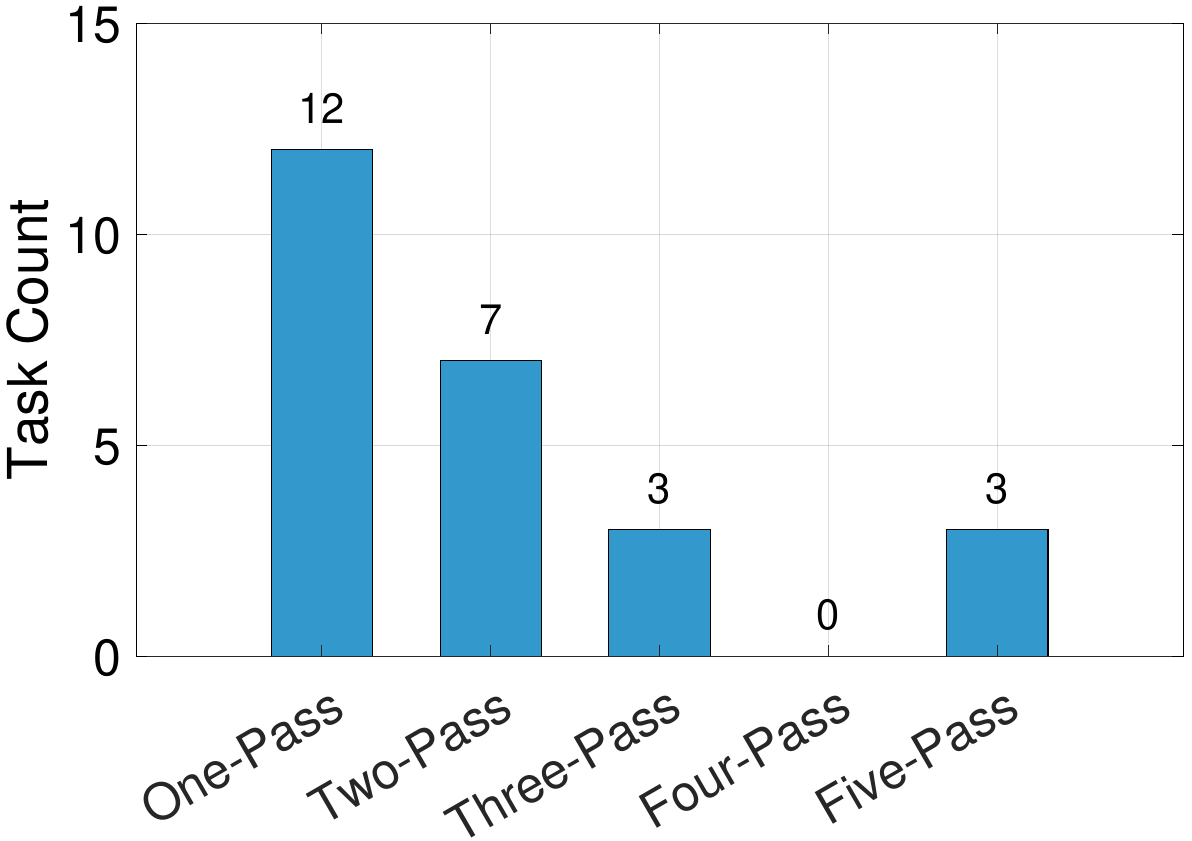}} 
    \caption{(a) Average passes for feedback-based fine-tuning across LLMs, (b) Tasks completed per pass in zero-shot.}
    \label{fig: tuning}
    \vspace{-10pt}
\end{figure}

As shown in Fig.~\ref{fig: tuning}(a), experiments across LLMs reveal that \fname{} generates robot control programs for user tasks within an average of two passes. To evaluate the effectiveness of feedback-based fine-tuning, we analyzed user tasks completed at each pass in a zero-shot setting. As depicted in Fig.~\ref{fig: tuning}(b), most tasks are solved within three passes, accounting for up to 88\% of the experimental tasks. These findings validate the efficiency and effectiveness of feedback-based fine-tuning, highlighting its applicability to real-world scenarios.

%% file: contents/7.conclusion.tex
\section{Conclusion}
\label{sec:conclusion}

This paper presents \fname{}, an LLM-powered natural-to-robotic language translation framework. The framework introduces \lname{} with its compiler to facilitate robot control program generation and provide correctness guarantees for generated programs. Additionally, a \lname{} debugger incorporates a feedback-based fine-tuning method to improve the success rate of program generation, enabling light-weight LLMs on resource-constrained devices. The experiments validate the effectiveness of the proposed framework and highlight its potential for intuitive error messaging and feedback-based fine-tuning. In future work, we will (i) integrate environmental feedback for real-world interactions; (ii) extend \lname{} to support more complex user tasks, and (iii) improve the fine-tuning process for \lname{} program generation.

\section{Acknowledgement}
\label{sec:acknowledge}

This work is supported by Guangzhou Research Funds under Grant 2023QN10X527.